\def\paperID{3085}
\def\confName{ICCV}
\def\confYear{2025}

\def\paperTitle{Interpretable Single-View 3D Gaussian Splatting using Unsupervised Hierarchical Disentangled Representation Learning}

\def\authorBlock{
    Yuyang Zhang\(^{1,2,3}\)\; \
    Baao Xie\(^{2,3}\)\thanks{Corresponding Author: Baao~Xie~\textless     bxie@idt.eitech.edu.cn\textgreater}\; \
    Hu Zhu\(^{2,3,4}\)\; \
    Qi Wang\(^{1,2,3}\)\; \
    Huanting Guo\(^{2,3}\)\; \
    Xin Jin\(^{2,3}\)\; \
    Wenjun Zeng\(^{2,3}\)\; \\
    $^1$Shanghai Jiao Tong University\\
    $^2$ Ningbo Institute of Digital Twin, Eastern Institute of Technology\\
    $^3$ Zhejiang Key Laboratory of Industrial Intelligence and Digital Twin,\\ Eastern Institute of Technology\\ 
    $^4$ Hong Kong Polytechnic University
    \\
}

\newif\ifreview 
\newif\ifarxiv 
\newif\ifcamera \newcommand{\cameraready}{\cameratrue}
\newif\ifrebuttal 

\cameraready %

\pdfoutput=1
\documentclass[10pt,twocolumn,letterpaper]{article}
\ifreview \usepackage[review]{cvpr} \fi
\ifarxiv \usepackage[pagenumbers]{cvpr} \fi
\ifrebuttal \usepackage[rebuttal]{cvpr} \fi
\ifcamera \usepackage{cvpr} \fi

\usepackage{graphicx}	
\usepackage{amsmath}	
\usepackage{amssymb}	
\usepackage{booktabs}
\usepackage{times}
\usepackage{microtype}
\usepackage{epsfig}
\usepackage{caption}
\usepackage{float}
\usepackage{placeins}
\usepackage{color, colortbl}
\usepackage{stfloats}
\usepackage{enumitem}
\usepackage{tabularx}
\usepackage{xstring}
\usepackage{multirow}
\usepackage{xspace}
\usepackage{url}
\usepackage{subcaption}
\usepackage{xcolor}
\usepackage[hang,flushmargin]{footmisc}
\usepackage{CJKutf8}
\usepackage{pifont}
\usepackage{indentfirst}
\usepackage{algorithm}
\usepackage{algpseudocode} 

\ifcamera \usepackage[accsupp]{axessibility} \fi

\definecolor{MyDarkBlue}{rgb}{0,0.5,1}
\definecolor{MyDarkGreen}{rgb}{0.02,0.6,0.02}
\definecolor{MyDarkRed}{rgb}{0.8,0.02,0.02}
\definecolor{MyDarkOrange}{rgb}{0.40,0.2,0.02}
\definecolor{MyYellow}{rgb}{1,0.55,0}
\definecolor{MyPurple}{RGB}{111,0,255}
\definecolor{MyRed}{rgb}{1.0,0.0,0.0}
\definecolor{MyGold}{rgb}{0.75,0.6,0.12}
\definecolor{MyDarkgray}{rgb}{0.66, 0.66, 0.66}
\definecolor{AliceBlue}{rgb}{0.61, 0.839, 0.89}
\definecolor{LightYellow}{HTML}{FFF8DC}
\definecolor{default}{RGB}{0,0,0}
\definecolor{tabfirst}{rgb}{1, 0.7, 0.7} %
\definecolor{tabsecond}{rgb}{1, 0.85, 0.7} %
\definecolor{tabthird}{rgb}{1, 1, 0.7} %

\newcommand{\cmark}{\ding{51}}
\newcommand{\xmark}{\ding{55}}

\ifarxiv  \fi

\newcommand{\R}[1]{{%
    \textbf{%
        \ifstrequal{#1}{1}{\textcolor{red}{R#1}}{%
        \ifstrequal{#1}{2}{\textcolor{blue}{R#1}}{%
        \ifstrequal{#1}{3}{\textcolor{magenta}{R#1}}{%
        \ifstrequal{#1}{4}{\textcolor{teal}{R#1}}{%
                           \textcolor{cyan}{R#1}%
        }}}}%
    }%
}}

\usepackage{xr-hyper}

\makeatletter
\newcommand*{\addFileDependency}[1]{
  \typeout{(#1)}
  \@addtofilelist{#1}
  \IfFileExists{#1}{}{\typeout{No file #1.}}
}

\makeatother
\newcommand*{\myexternaldocument}[1]{
    \externaldocument{#1}
    \addFileDependency{#1.tex}
    \addFileDependency{#1.aux}
}

\definecolor{cvprblue}{rgb}{0.21,0.49,0.74}
\usepackage[pagebackref,breaklinks,colorlinks,allcolors=cvprblue]{hyperref}
\usepackage[capitalize]{cleveref}
\crefname{section}{Sec.}{Secs.}
\crefname{table}{Table}{Tables}
\crefname{figure}{Fig.}{Figs.}

\ifarxiv \crefname{appendix}{App.}{Apps.}
\else \crefname{appendix}{Suppl.}{Suppls.} \fi

\unless\ifarxiv \myexternaldocument{_supplementary} \fi

\begin{document}
\title{\paperTitle}
\author{\authorBlock}
\maketitle

\begin{abstract}
    Gaussian Splatting (GS) has recently marked a significant advancement in 3D reconstruction, delivering both rapid rendering and high-quality results. However, existing 3DGS methods pose challenges in understanding underlying 3D semantics, which hinders model controllability and interpretability. To address it, we propose an interpretable single-view 3DGS framework, termed 3DisGS, to discover both coarse- and fine-grained 3D semantics via hierarchical disentangled representation learning (DRL). Specifically, the model employs a dual-branch architecture, consisting of a point cloud initialization branch and a triplane-Gaussian generation branch, to achieve coarse-grained disentanglement by separating 3D geometry and visual appearance features. Subsequently, fine-grained semantic representations within each modality are further discovered through DRL-based encoder-adapters. To our knowledge, this is the first work to achieve unsupervised interpretable 3DGS. Evaluations indicate that our model achieves 3D disentanglement while preserving high-quality and rapid reconstruction.
\end{abstract}

\section{Introduction}
\begin{figure}
    \centering
    \subfloat[Conventional 3DGS Reconstruction]{\includegraphics[width=\linewidth]{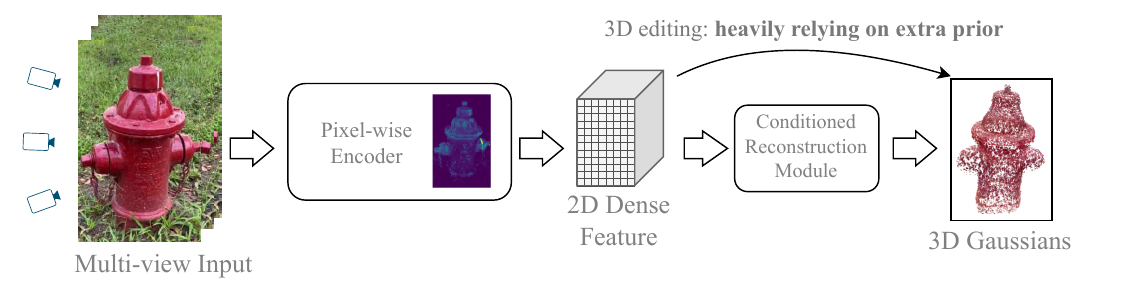}}\\
    \subfloat[3DisGS: \textbf{Interpretable Single-view 3DGS Reconstruction}]{\includegraphics[width=\linewidth]{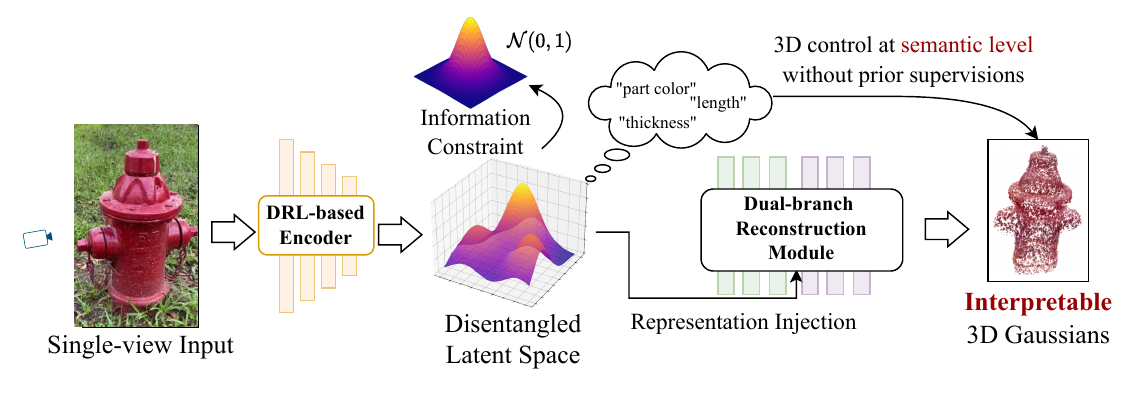}}
    \vspace{-2mm}
    \caption{The comparison of (a) conventional 3DGS and (b) proposed 3DisGS. Traditional models are inherently non-interpretable, limiting 3D editing to pixel-level and relying heavily on extra priors (masks, bounding boxes, \etc). In contrast, 3DisGS employs hierarchical DRL to achieve interpretable 3D reconstruction unsupervisedly, which enables attribute manipulation at semantic-level.}
    \label{fig:teaser}
\end{figure}
Despite advancements of implicit vision-based 3D reconstruction (V3DR) technologies including Neural Radiance Fields (NeRF)~\cite{mildenhall2021nerf} and Signed Distance Functions (SDF)~\cite{park2019deepsdf}, these techniques encounter constraints in terms of computational efficiency and controllability against implicit-explicit approaches like 3D Gaussian Splatting (3DGS). Specifically, 3DGS reconstructs 3D scenes via adaptive anisotropic Gaussians optimized from Structure from motion (SfM) points, dynamically refining density and rendering through splatting for real-time view synthesis without mesh/voxel representations~\cite{kerbl20233d}. Based on this, extensive efforts have been made to enhance 3DGS in terms of quality, speed, and optimization, resulting in the family of 3DGS-based approaches~\cite{zhou2024feature,yu2024gsdf,qian20243dgs}.

However, a fundamental challenge remains for the 3DGS-based approaches and, as well as for all learning-based \textit{``black-box"} 3D models: the limited interpretability inherent in neural representations~\cite{xie2023navinerf}. This limitation implies that current approaches struggle to discover and identify latent semantics behind the 3D observations as biological intelligence does. For instance, while a 3DGS model can reconstruct an indoor scene, it lacks a fundamental understanding of 3D semantic concepts related to Gaussian ellipsoids, such as ``furniture", ``decorations", ``persons" and \etc, let alone control and edit even more fine-grained concepts. Disentangled representation learning (DRL) is developed to addresses such interpretability challenges by imitating the understanding processes of biological intelligence, which decompose observations into independent factors~\cite{xie2025graph}. This enables specific attributes (e.g., color, shape, size) to respond exclusively to changes in corresponding factors. While extensively studied in 2D settings, DRL remains underexplored in 3D scenes due to the complexity and topology of 3D environments.

To address this challenge, we propose 3DisGS, an unsupervised interpretable 3D reconstruction framework achieves both coarse- and fine-grained 3D disentanglement through a hierarchical DRL architecture. Specifically, our model comprises two key components: a dual-branch reconstruction module and DRL-based encoder-adapters, each responsible for the disentanglement at coarse- and fine-level, respectively. The reconstruction module comprises two synergistic branches for geometry and appearance reconstruction. The point cloud initialization branch (referred as ``geometry branch") adopts a folding-based decoder to deform 2D grid primitives into a set of initial 3D points, while the triplane-Gaussian generation branch (referred as  ``appearance branch") leverages these points to build a locally continuous Gaussian triplane.

Following the coarse-grained disentanglement, DRL-based encoder-adapters are designed to unsupervisedly extract the disentangled semantic factors. Specifically, given a 2D image input, the model utilizes a pretrained ViT backbone (DINOv2)~\cite{caron2021emerging} to extract high-level 2D features, which are subsequently processed by dual convolutional encoders. Each encoder independently transmits the encoded features to its corresponding DRL adapter. By enforcing DRL constraints, the encoder-adapters construct an orthogonal latent space, with each dimension encoding distinct and interpretable semantic factors. This design facilitates fine-grained disentanglement of geometry and appearance in an independent manner. To ensure effective reconstruction with such compact conditions, specific style-guided modules are tailored for each branch. Furthermore, a mutual information loss is introduced to reduce appearance overfitting by facilitating the transfer of 3D structural information between the appearance latent space. In summary, our contributions are:
\begin{enumerate}
    \item To the best of our knowledge, the proposed 3DisGS is the first 3DGS-based interpretable reconstruction framework that utilizes only single-view inputs, without additional supervision.
    \item The proposed dual-branch framework leverages a hierarchical DRL strategy to achieve coarse-to-fine disentanglement for both the 3D geometry and visual appearance.
    \item To ensure view-consistent 3D reconstruction from single-view inputs, we employ style-guided modules and mutual information loss to enhance the 3D information extraction and transformation.
\end{enumerate}
Experimental results demonstrate the effectiveness of the proposed approach in 3D disentanglement across both synthetic and real-world datasets, while maintaining high reconstruction quality and computational efficiency.

\section{Related Works}
\subsection{3D Gaussian Splatting}

3D Gaussian Splatting (3DGS) has recently emerged as a transformative technique in 3D reconstruction domain. This approach, characterized by the utilization of millions of 3D Gaussians, represents a significant departure from NeRF-based methods~\cite{wu2024recent} that predominantly rely on implicit models to map spatial coordinates to pixel values. Specifically, 3DGS leverages a set of parameterized 3D Gaussians, each defined by its spatial position, covariance matrix, and associated attributes such as color and opacity~\cite{kerbl20233d}. These Gaussians are projected onto the image plane via a splatting process, enabling efficient and continuous rendering of complex scenes. By adopting explicit representations, 3DGS achieves superior rendering speed and scalability, making it particularly suitable for several tasks like dynamic reconstruction~\cite{guo2024motion,fan2024spectromotion,li2024occscene}, virtual reality (VR)~\cite{kim20243dgs,qiu2024advancing,meng2024mirror}, augmented reality (AR)~\cite{li2024robogsim,du2024mvgs}, digital twins~\cite{do2024hologaussian,wang2025scene,wang2024tortho}. However, Current 3DGS methods face limitations in 3D semantic perception, leading to reduced controllability and generalizability.

\subsection{Disentangled Representation Learning}
Disentangled Representation Learning (DRL) was intuitively introduced by Bengio et al.~\cite{bengio2013representation} as a paradigm aimed at enhancing interpretability by decomposing the semantic factors underlying observational data~\cite{jin2024closed}. This approach assumes that specific attributes are sensitive to changes in single latent factors, while not being affected by others. Currently, unsupervised DRL methods primarily utilize the Variational Autoencoder (VAE)~\cite{kingma2013auto}, a probabilistic generative model that learns disentangled representations through the incorporation of a Kullback-Leibler divergence term. This framework has been further refined and extended by models such as $\beta$-VAE~\cite{higgins2017beta}, $\beta$-TCVAE~\cite{chen2018isolating}, FactorVAE~\cite{kim2018disentangling}, and $\alpha$-TCVAE~\cite{meo2023alpha} via improvements in regularization techniques. Despite these advancements, limited research has explored the integrations of DRL in the 3D domain, where semantic-aware representation learning is of critical importance.

\subsection{Interpretable 3D Reconstruction}
Existing 3D disentanglement approaches primarily focus on the separation of geometry and appearance. For example, Tewari et al.~\cite{tewari2022disentangled3d} introduced a NeRF-GAN framework capable of disentangling geometry, appearance, and camera pose from monocular images. Furthermore, Chen et al.~\cite{chen2023fantasia3d} present a novel approach for high-quality text-to-3D generation, which disentangles geometry and appearance modeling to achieve accurate geometry reconstruction and photorealistic per-view rendering. Xu et al.~\cite{xu2024texture} propose a 3DGS-based model that extracts 3D appearance information by representing it as a 2D texture mapped onto the 3D surface, to enable more flexible 3D editing. However, the form of ``disentanglement" employed in existing works primarily focuses on explicit attributes (\ie geometry and appearance), while overlooking the disentanglement of more abstract and semantic latent representations. Consequently, these methods are limited in their ability to enable models to learn and understand the semantic concepts of reconstructed scenes.

\section{Preliminary}
\subsection{3D Gaussian Splatting}
3DGS is a volumetric scene representation that models a 3D scene as a collection of anisotropic Gaussian primitives. Formally, each primitive is parameterized by its position $\mathbf{\mu} \in \mathbb{R}^3$, covariance matrix $\mathbf{\Sigma} \in \mathbb{R}^{3\times3}$, color $\mathbf{c} \in \mathbb{R}^3$ encoded with spherical harmonics (SHs) and opacity $\alpha \in [0,1]$. The 3D scene is thus represented as a mixture of Gaussians: $\mathcal{G} = \{(\mathbf{\mu}_i, \mathbf{\Sigma}_i, \mathbf{c}_i, \alpha_i)\}_{i=1}^N$. Building on this, the subsequent rendering process follows a volumetric paradigm. For a camera ray $\mathbf{r}(t) = \mathbf{o} + t\mathbf{d}$, classical volume rendering computes pixel color by integrating radiance along the ray:  
$$
\mathbf{C}(\mathbf{r}) = \int_{t_n}^{t_f} T(t) \sigma(\mathbf{r}(t)) \mathbf{c}(\mathbf{r}(t), \mathbf{d}) \, dt,
$$
where $ T(t) = \exp\left(-\int_{t_n}^t \sigma(\mathbf{r}(s)) \, ds\right)$ is transmittance, and $\sigma$ denotes density. In the 3DGS framework, this continuous integral is approximated by discretizing the scene into a set of overlapping Gaussians. Each Gaussian contributes a density $\sigma_i = \alpha_i \mathcal{G}_i(\mathbf{x})$, where $\mathcal{G}_i$ is the anisotropic 3D Gaussian kernel. 

During the rasterization, 3D Gaussians are projected to image space via perspective projection. The projected 2D covariance, denoted as $\mathbf{\Sigma'}$, can be computed as follows:  
$$
\mathbf{\Sigma'} = \mathbf{J W \Sigma W^\top J^\top}
$$  
where $\mathbf{W}$ is the viewing transform and $\mathbf{J}$ is the Jacobian of the affine approximation. The final pixel color aggregates contributions from $K$ depth-ordered Gaussians through alpha compositing:  
$$
\mathbf{C} = \sum_{i=1}^K \mathbf{c}_i \alpha_i \prod_{j=1}^{i-1} (1 - \alpha_j)
$$
This formulation maintains differentiability, enabling joint the optimization of Gaussian parameters ($\mathbf{\mu}, \mathbf{\Sigma}, \mathbf{c}, \alpha$) via gradient descent on a photometric loss. In contrast to implicit volumetric representations, 3DGS achieves real-time rendering by leveraging GPU-accelerated tile-based splatting, while maintaining high fidelity through the use of adaptive anisotropic Gaussians. Furthermore, the discrete nature of 3DGS provides a more suitable framework for interpretable and controllable 3D reconstruction compared to purely implicit representations.

\section{Methodology}

\begin{figure*}[htbp]
  \centering
  \includegraphics[width=\linewidth]{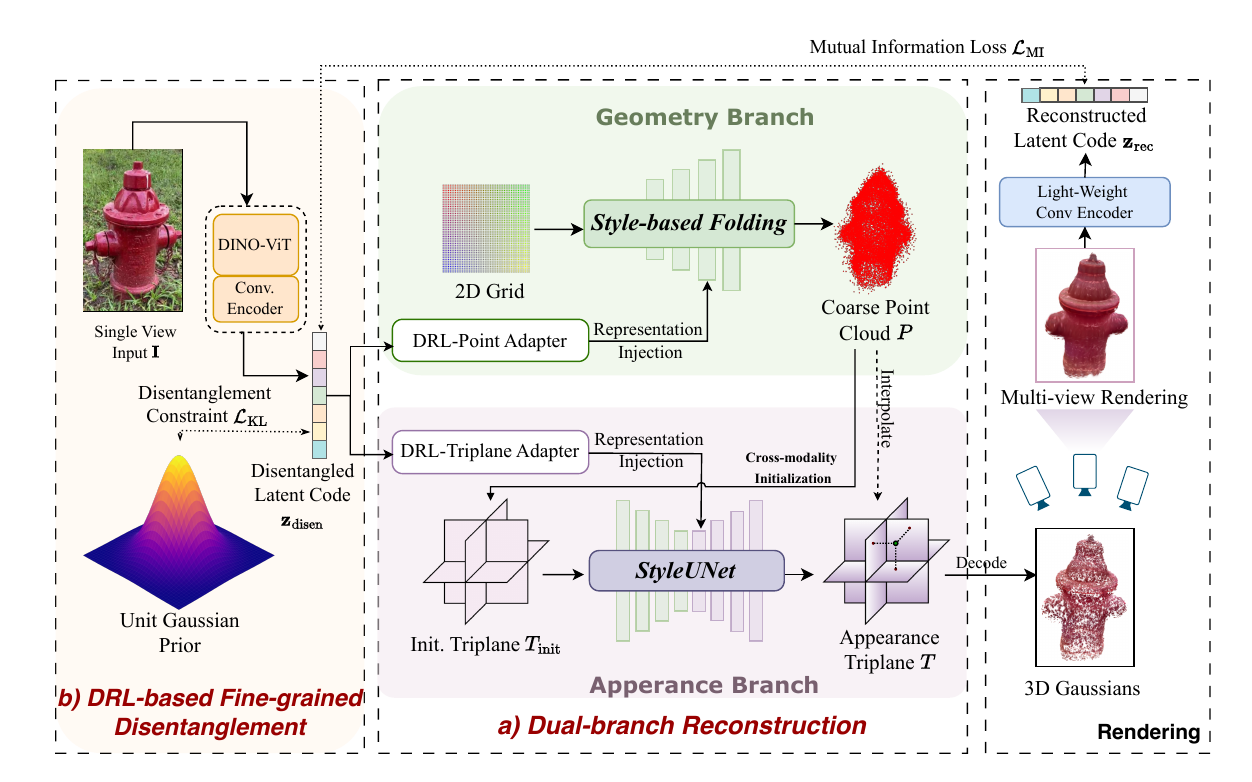}
  \vspace{-8mm}
  \caption{The overview of 3DisGS. Given a single-view image \(I\), a pretrained DINO-ViT is employed to extract rich features, which are subsequently compressed into compact, disentangled latent code~\( z_{disen}\) via DRL-based encoder. This interpretable code is adapted by DRL-based adapters to modality-specific forms and fed to two branches. The geometry branch generates point clouds, serving as the initialization for appearance branch to produce a triplane \(T_{\text{init}}\). The triplane features are then decoded into 3D Gaussians. To improve reconstruction and  disentanglement, a mutual information loss~\(\mathcal{L}_{\text{MI}}\) is applied among~\( z_{disen}\) and the reconstructed outputs.}
  \label{fig:arc}
\end{figure*}

As depicted in Figure~\ref{fig:arc}, we present the details of 3DisGS, which involves three main components:
\begin{enumerate}[label=\alph*)]
  \item Dual-branch Reconstruction Module: consists of two synergistic reconstruction branches, each integrated with style-guided modules, to achieve coarse-grained disentanglement on geometry and appearance.~(\cref{sect:coarse}).
  \item DRL-based Encoder-Adapter: extracts disentangled representations from image and adapt them for reconstruction. ~(\cref{sect:disentangle}).
  \item Loss Functions: includes the specifically designed mutual information loss and other loss functions incorporated in the optimization process.~(\cref{sect:loss}).
\end{enumerate}

\subsection{Dual-branch Reconstruction}
\label{sect:coarse}
Typical 3DGS models use a feed-forward process with a single-stage module mapping pixels to 3D Gaussians. However, the reliance on low-level positional information due to the per-pixel alignment nature, leads to the entanglement of 3D geometry and visual appearance. Towards it, we propose a dual-branch framework that separates reconstruction into point cloud based 3D geometry and triplane-based visual appearance. This design enables a coarse-grained disentanglement between geometry and appearance while facilitating a progressive reconstruction process with improved quality.

In this section, we provide a detailed reconstruction process for each branch, guided by the assumed disentangled latent representations \(\mathbf{c}_{\text{apr}}, \mathbf{c}_{\text{pcd}}\). Given these conditions, the geometry branch $\text{Geo}$ reconstructs the point cloud \(P\in \mathbb{R}^{N\times 3}\), while the appearance branch $\text{Apr}$ generates triplane feature \(T\in \mathbb{R}^{3\times N_p \times N_p \times C_p}\). The reconstruction process is formally expressed as follows:
\[\begin{aligned}
  P &= \text{Geo}(\mathbf{c}_{\text{pcd}}), \\
  T &= \text{Apr}(P, \mathbf{c}_{\text{apr}}).\\
\end{aligned}
\]
where the overall reconstruction module can be subsequently defined as: \( R = \text{GaussianDec}(P, T)\)

\subsubsection{Geometry Reconstruction Branch}
\label{sect:pcd}
As illustrated by the green block in Figure~\ref{fig:arc}, we integrate a style-based folding module in the geometry reconstruction branch. By combining folding~\cite{yang2018foldingnet} with style-based representation injection, our model enables semantically disentangled and hierarchical control over generated point cloud. Further, it excels in representing complex structures using 1D latent code, aligning well for subsequent DRL-based adaptations. 

Given \(N\) initial grid primitives from unit square~\([0,1]^2\), denoted as $P_{\text{init}} \in \mathbb{R}^{N \times Res \times 2}$, the style-based folding module employs a mapping function \(F_{\text{fold}}\), implemented as an multilayer perceptron (MLP), to transform 2D grid into 3D point cloud. This folding process can be formulated as:
\[
P = F_{\text{fold}}(P_{\text{init}}, \mathbf{c}_{\text{pcd}}),
\]
where \(P \in \mathbb{R}^{N \times 3}\) represents the generated 3D point cloud. The transformation is guided by the latent code \(\mathbf{c}_{\text{pcd}}\), which is derived through the DRL-based encoder-adapter via representation injection. Specifically, the representation injection transforms the batch-normalized 
intermediate feature \( \bar{\mathbf{h}}_{\text{in}}\) of MLP into the stylized feature \( \mathbf{h}_{\text{out}}\) with the latent code \(\mathbf{c}_{\text{pcd}}\):
\[
\mathbf{h}_{\text{out}} = \gamma_{\mathbf{c}} \odot \bar{\mathbf{h}}_{\text{in}} + \beta_{\mathbf{c}},
\]
where \(\gamma_{\mathbf{c}}, \beta_{\mathbf{c}}\) are modulation parameters derived from \(\mathbf{c}_{\text{pcd}}\). The geometry branch establishes the foundational structure of the 3D model, serving as the skeleton and initialization for subsequent appearance reconstruction.

\subsubsection{Appearance Reconstruction Branch}
\label{sect:apr}
The appearance branch generates the feature triplane \(T\), used for interpolating Gaussian features and determining key attributes, including refined positions, rotations, SHs, and opacity, which define the model's detailed appearance.

To achieve effective disentanglement between appearance and geometry, it is crucial for the appearance branch to be predominantly influenced by the appearance condition \(\textbf{c}_{\text{apr}}\), as provided by the DRL-Triplane adapter. Nonetheless, incorporating geometric information remains indispensable to maintain consistency and alignment between the two reconstruction modalities. To address this challenge, we introduce a style-based U-Net, termed StyleUNet, which separately delivers geometry and appearance representations to the triplane. As illustrated by the red block in  Figure~\ref{fig:arc}, given the reconstructed point cloud \(P\) as condition, the appearance branch encodes local-geometry feature~\(\textbf{F}_{\text{local}}\) with a local-pooled PointNet~\cite{peng2020convolutional, qi2017pointnet} and subsequently projected onto the triplane to initialize the feature representation:
\[
  \begin{aligned}
  T_{\text{init}} &=\{T_{XY},T_{XZ},T_{YZ}\}= \text{Proj}(\textbf{F}_{\text{local}}, P),\\
  \end{aligned}
\]
where \(T_{\text{init}} \in \mathbb{R}^{3 \times N_p \times N_p \times C_p}\) represents the initial triplane features that embed local geometry information. The axial projection~\(\text{Proj}(\textbf{F}_{\text{local}}, p)\) performs mean pooling on the point features along each axis. 

To ensure a balanced integration of information between branches, the initial triplane features are further encoded using the encoders of StyleUNet~\(\{\text{Enc}_i\}\). The encoders compress the features into lower-resolution representations~\(\textbf{FC}_{i}\) with reduced channel dimensions, thereby refining the information for subsequent processing. During the decoding process, starting from the lowest-level of StyleUNet feature \(\textbf{FC}_{\text{low}} = \textbf{FC}_ {i_\text{max}}\), the triplane features~\(\textbf{FR}_i\) are progressively reconstructed using the StyleUNet decoders~\(\{\text{Dec}_i\}\). This reconstruction integrates the previously encoded geometric information and incorporates the appearance condition 
through representation injection, as defined by:
\[
  \textbf{FR}_{i} = \text{Dec}_i(\textbf{FR}_{i+1}, \textbf{FC}_{i},\textbf{c}_{\text{apr}}), \quad i = i_{\text{max}-1} \to 0.
\]
where the decoder operation is expressed as~\(\text{Dec}_i = \text{StyleConv}(\text{Comb}(\textbf{FR}_{i+1}, \textbf{FC}_{i}),\textbf{c}_{\text{apr}})\). This decoding process constrains the complexity of geometric encoding, ensuring a balanced integration of information from different modalities and reducing the risk of overfitting. The final triplane feature~\(T=\textbf{FR}_0\) is then sampled using the point cloud~\(P\) through bilinear interpolation:
\begin{align*}
  p_{XY} = (x, y),
\mathbf{f}_{XY} = \text{Interp}(p_{XY}, \mathbf{T}_{XY}),
\end{align*}
where the same interpolation process is applied to the \(XZ, YZ\) plane. The interpolated triplane features~\(\mathbf{f}_{XY}\) are concatenated to form the final feature~\(\mathbf{f}\):
\[
\mathbf{f}_p = \mathbf{f}_{XY}\oplus \mathbf{f}_{XZ}\oplus \mathbf{f}_{YZ}.
\]
Finally, ~\(\mathbf{f}_p\) is passed through a shallow MLP to obtain attributes~\((\mathbf{\mu}_i, \mathbf{\Sigma}_i, \mathbf{c}_i, \alpha_i)\) of individual Gaussian primitives.

\subsection{DRL-based Fine-grained
Disentanglement}
\label{sect:disentangle}
To achieve fine-grained representation disentanglement in both 3D geometry and visual appearance, DRL-based encoder-adapters (\ie DRL-Point Adapter and DRL-Triplane Adapter) are proposed to extract interpretable semantics.

As illustrated by the yellow block in Figure~\ref{fig:arc}, given a single-view input image \( I \in \mathbb{R}^{ H\times W\times 3}\), the model first extract rich features \( \mathbf{F}_I \in \mathbb{R}^{H_p \times W_p \times C} \) with a pretrained ViT backbone (DINOv2). These features \( \mathbf{F}_I \) processed through a convolutional encoder for channel and spatial compression, yielding a reduced representation \( \mathbf{F}_{\text{comp}} \in \mathbb{R}^{H' \times W' \times C'} \). The compressed feature \( \mathbf{F}_{\text{comp}} \) is flattened into a 1D vector, passed through an MLP for further dimensionality reduction, which parameterized as the mean \(\mu\) and variance \(\sigma\) of a posterior Gaussian distribution. Subsequently, a low-dimensional latent code \( \mathbf{z} \in \mathbb{R}^{d} \) (where \( d \ll C' \times H' \times W' \)) is sampled, encapsulating the disentangled semantic factors. This disentangled latent code is then transformed via different adapters into conditioning representations \( \mathbf{c}_{\text{apr}}, \mathbf{c}_{\text{pcd}} \in \mathbb{R}^{C_{con}} \) or \(\mathbb{R}^{H_p\times W_p\times C_{con}}\) , which are compatible for geometry and appearance branch, respectively.

While the processes above compresses the input image into a compact latent code, it does not inherently promise disentanglement. To address this, we impose latent space constraints on the encoder-adapter, guided by principles of $\beta$-VAE~\cite{higgins2017beta} and information bottleneck theory. Specifically, $\beta$-VAE learns latent representations of observations by approximating data distribution via a maximum likelihood estimation:
\begin{equation}\label{eq1}
  \log p_{\theta}(\mathbf{x})=D_{K L}\left(q_{\phi}(\mathbf{z}|\mathbf{x}) \| p_{\theta}(\mathbf{z}|\mathbf{x})\right)+\mathcal{L}(\theta, \phi),\\
\end{equation}
where $q_{\phi}(\mathbf{z}|\mathbf{x})$ is the estimated posterior distribution of latent $\mathbf{z}$ given observation $\mathbf{x}$. The optimization objective of \cref{eq1} is to maximize the evidence lower bound $\mathcal{L}(\theta, \phi)$. This goal can be decomposed into two parts as:
\begin{equation}\label{eq2}
  \mathcal{L}(\theta, \phi)=\mathbb{E}_{q_{\phi}(\mathbf{z}|\mathbf{x})}\left[\log p_{\theta}(\mathbf{x} | \mathbf{z})\right]-\beta D_{K L}\left(q_{\phi}(\mathbf{z} | \mathbf{x}) \| p(\mathbf{z})\right),
\end{equation}
where the initial term, is responsible for the reconstruction quality, and the second term, \ie, KL divergence $D_{K L}\left(q_{\phi}(\mathbf{z} | \mathbf{x}) \| p(\mathbf{z})\right)$,  constraints the latent space to be close to a prior distribution $p(\mathbf{z})$. To improve disentanglement, $\beta$-VAE based models introduce an explicit inductive bias through hyperparameter $\beta$ of the KL term. The $\beta$ penalty intensifies the independence constraint on posterior distribution, thereby enhancing the model's ability to separate underlying factors of variation in the data. 

\subsection{Loss Function}
\label{sect:loss}
\subsubsection{Mutual Information Loss}
Besides DRL constraints, a mutual information loss is introduced to enhance branch disentanglement by maximizing the mutual information between the disentangled latent code and 2D rendered views. Denoting the overall reconstruction module, including the adapter, as \(\text{Rec}(\mathbf{z}_{\text{apr}}, \mathbf{z}_{\text{pcd}})\), the mutual information loss is defined as:
\begin{align*}
\mathcal{L}_{\text{MI}} &= I(\mathbf{z}_\text{apr};\text{LightEnc}(\text{Rec}(\mathbf{z}_{\text{apr}}, \mathbf{z}_{\text{pcd}}))),
\end{align*}
This term can be reformulated as the likelihood between the estimated posterior distribution \(p(\mathbf{z}_{\text{apr}}|\mathbf{x})\) and the decoded latent code derived from the 2D renderings. With \(\mathcal{L}_{\text{MI}}\), the overall DRL constraints can be defined as:
\begin{align*}
  \mathcal{L}_{\text{DRL}} &= \mathcal{L}_{\text{KL}} + \mathcal{L}_{\text{MI}}\\
  &= \beta D_{KL}\left(q_{\phi}(\mathbf{z} | \mathbf{x}) \| p(\mathbf{z})\right) + \\ &\quad\alpha I(\mathbf{z}_\text{apr};\text{LightEnc}(\text{Rec}(\mathbf{z}_{\text{apr}}, \mathbf{z}_{\text{pcd}}))),
\end{align*}

\subsubsection{Reconstruction Loss}
To optimize the reconstruction branches, we design separate loss functions tailored to the geometry and appearance reconstruction tasks. For the geometry reconstruction branch, we employ a point cloud reconstruction loss \(\mathcal{L}_{\text{pc}}\), which is computed using the Earth Mover's Distance (EMD) between the predicted point cloud \(\mathbf{P}_{\text{pred}}\) and the ground truth \(\mathbf{P}_{\text{gt}}\):
\[
\mathcal{L}_{\text{pc}} = \text{EMD}(\mathbf{P}_{\text{pred}}, \mathbf{P}_{\text{gt}}).
\]
For the appearance reconstruction branch, we utilize a Gaussian decoder and define a rendering loss~\(\mathcal{L}_{\text{render}}\) to capture both pixel-level and perceptual differences. The loss combines Mean Squared Error (MSE), Structural Similarity Index (SSIM), and Learned Perceptual Image Patch Similarity (LPIPS), computed between the rendered image \(\mathbf{I}_{\text{pred}}\) and the ground truth \(\mathbf{I}_{\text{gt}}\) across a batch of \(N\) images:
\[
\mathcal{L}_{\text{render}} = \sum_{i=1}^N (\lambda_{\text{m}} \mathcal{L}_{\text{MSE}} + \lambda_{\text{s}} \mathcal{L}_{\text{SSIM}} + \lambda_{\text{l}} \mathcal{L}_{\text{LPIPS}}
)+\lambda_{\text{reg}} \mathcal{L}_{\text{reg}}.\]
To further enhance appearance fidelity and prevent overfitting, a regularization term \(\mathcal{L}_{\text{reg}}\) is added. This term includes L1 Loss, which enforces sparsity, and Total Variation (TV) Loss, promoting smoothness in the rendered images:
\[
\mathcal{L}_{\text{reg}} = \lambda_{\text{L1}} \mathcal{L}_{\text{L1}} + \lambda_{\text{TV}} \mathcal{L}_{\text{TV}}.
\]

\subsubsection{Total Loss}
The overall training objective of the proposed model is formulated as a composite loss function \(\mathcal{L}\):
\[
\mathcal{L} = \mathcal{L}_{\text{recon}} + \mathcal{L}_{\text{DRL}},
\]
where \(\mathcal{L}_{\text{recon}}\) represents the reconstruction loss and \(\mathcal{L}_{\text{DRL}}\) is the DRL-based constraints. This unified objective ensures robust reconstruction of both 3D geometry and visual appearance through the synergy of these tailored loss functions.

\begin{algorithm*}[t]
    \caption{The training pipeline of 3DisGS.}
    \label{algo:overall}
    \begin{algorithmic}[1]
    \small
    \State \textbf{Require:} Dataset $\mathcal{D} = \{(\mathbf{I_{\text{in}}}_i, \{\mathbf{I}_\text{novel}\}_i, \mathbf{P}_i)\}_{i=1}^N$, where $\mathbf{I_{\text{in}}}_i$ and $\{\mathbf{I}_\text{novel}\}_i$ are posed input, output images with point cloud $\mathbf{P}_i$.
    \State \textbf{Initialize:} The parameter \(\{\phi,\theta,\gamma,\tau,\alpha,\beta,\xi,\mathcal{O}\}\) of model, include posterior encoders $q_\phi(\mathbf{z}|\mathbf{I}), q_\theta(\mathbf{z}|\mathbf{I})$, point cloud, appearance reconstruction module $\text{Geo}_\tau$, $\text{Apr}_\gamma$, lightweight encoder \(\text{LightEnc}_\xi\), optimizer $\mathcal{O}$.
    \For{epoch $=1, \cdots, N$}
        \For{each batch $(\mathbf{I}_\text{in},\mathbf{P}_\text{in},\{\mathbf{I}_\text{novel}\} )\in \mathcal{D}$}
            \State Encode image to posterior \(q_{\phi}(\mathbf{z}_{apr}|\mathbf{I}_{\text{in}})\) and \(q_{\theta}(\mathbf{z}_{pcd}|\mathbf{I}_{\text{in}})\).
            \Comment{DRL-based Encoder-Adapter(\cref{sect:disentangle})}
            \State Sample \(z_{apr} \sim q_{\phi}(\mathbf{z}_{apr}|\mathbf{I}_{\text{in}})\) and \(z_{pcd} \sim q_{\theta}(\mathbf{z}_{pcd}|\mathbf{I}_{\text{in}})\), transform to condition \(\textbf{c}_{\text{apr}},\textbf{c}_{\text{pcd}}\).
            \State Reconstruct \(\hat{P}=\text{Geo}(\mathbf{c}_{\text{pcd}})\).
            \Comment{Geometry(\cref{sect:pcd})}
            \State Reconstruct triplane \(T=\text{Apr}_\gamma(\hat{P},\textbf{c}_\text{apr})\), Interpolate $T$ with $\hat{P}$, decoding to Gaussians.
            \Comment{Appearance(\cref{sect:apr})}
            \State Render to novel view $\{\hat{I}_\text{novel}\}$
            \State Compute DRL loss and reconstruction loss, update the model parameters with $\mathcal{O}$.
            \Comment{Losses(\cref{sect:loss})}
        \EndFor
    \EndFor
    \end{algorithmic}
\end{algorithm*}

\section{Experiments}
\subsection{Datasets}
We evaluate our model on standard benchmarks: 1)~\textbf{ShapeNet Chairs}~\cite{chang2015shapenet}, comprising over 5,000 3D CAD models of chairs; 2)~\textbf{ShapeNet Cars} featuring more than 3,000 3D CAD models of cars; 3)~\textbf{ShapeNet Airplane}, containing over 3000 models of airplanes. 4)~\textbf{CO3D Hydrant}~\cite{reizenstein2021common}, which includes over 300 capture sequences of real-world hydrants. For details on dataset initialization, please refer to the appendix.

\subsection{Implementation Details}
In all experiments, the latent code \(z\) is set to a dimension of 32. The model is trained using the Adam optimizer with a learning rate of 6e-5 and a batch size of 32, scheduled via a warm-up cosine annealing strategy with one warm-up epoch. All experiments were conducted on 4 NVIDIA A800 80G GPUs using PyTorch 2.0.0 and CUDA 11.7.

\begin{figure*}[ht]
  \centering
  \subfloat[Disentanglement results on 3D geometry.]{\includegraphics[width=0.6\linewidth]{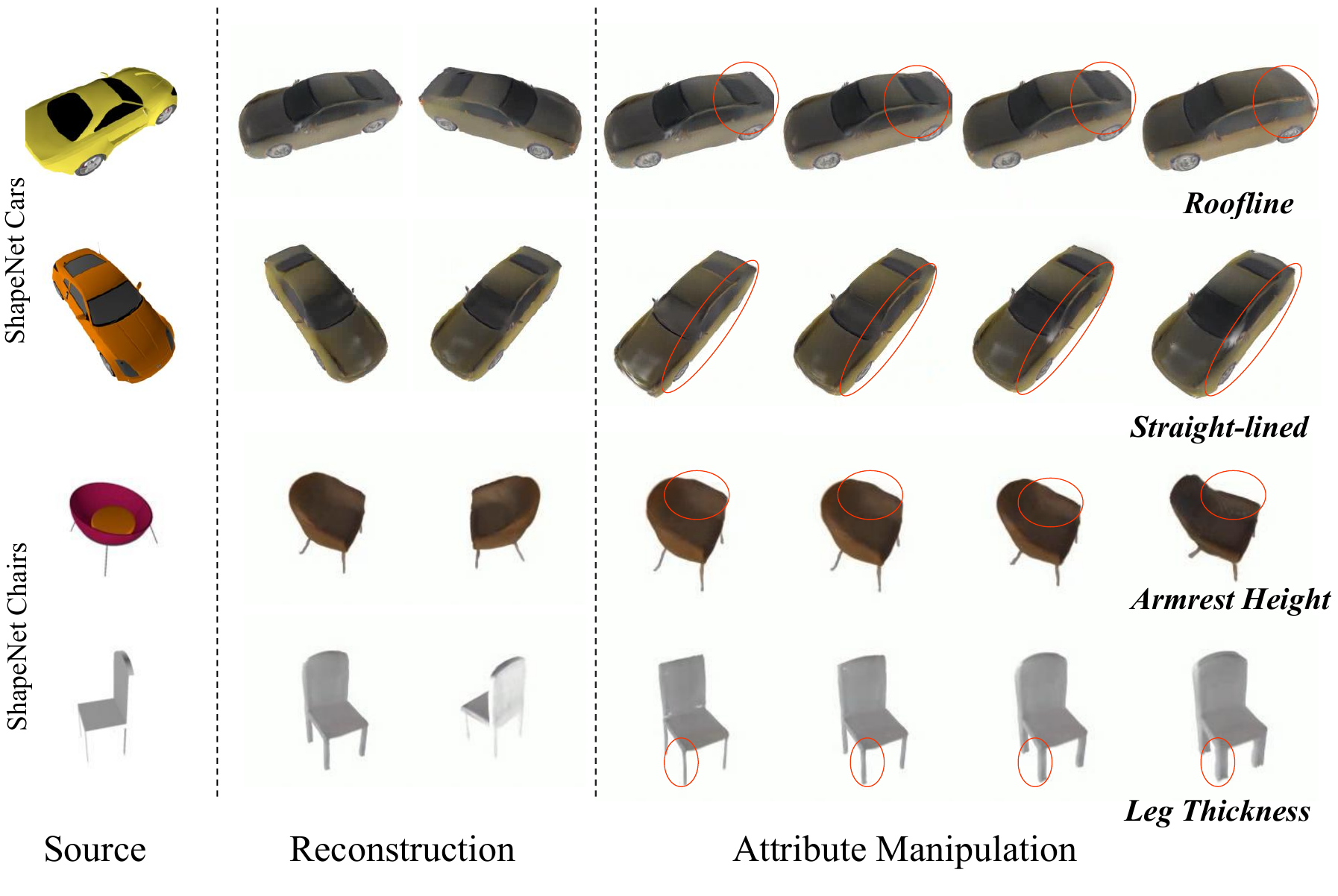}}
  \subfloat[Disentanglement results on visual appearance.]{\includegraphics[width=0.4\linewidth]{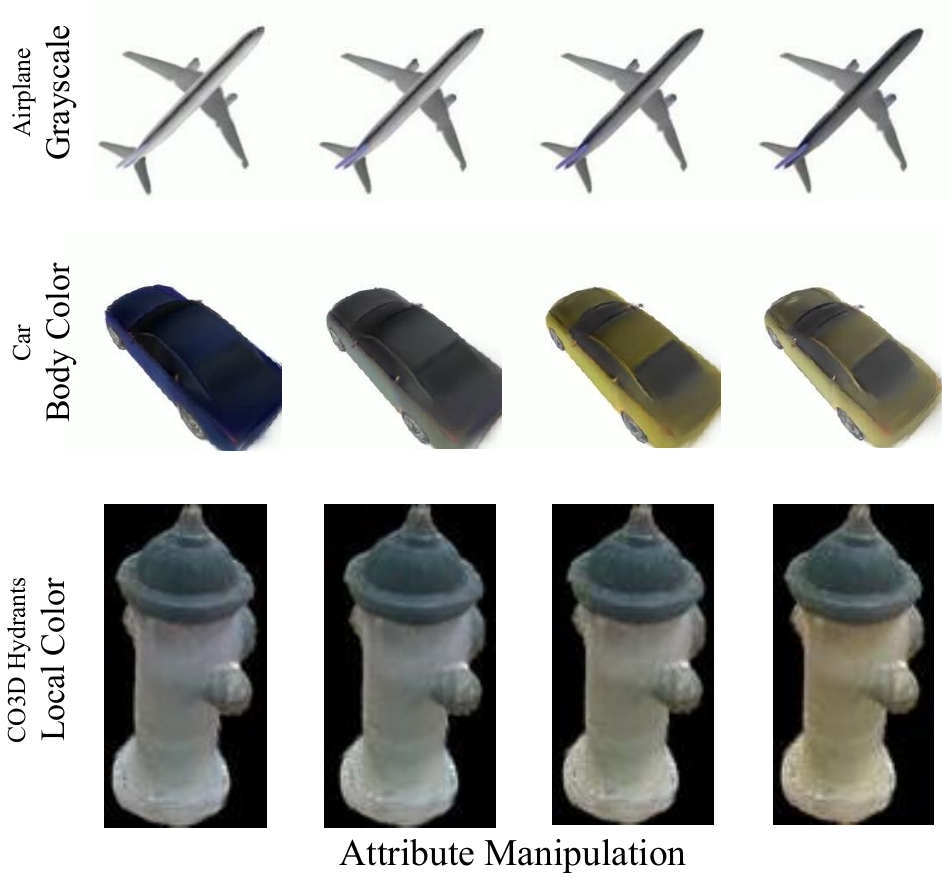}}
    \vspace{-1mm}
  \caption{\textbf{Interpretable 3D reconstruction results.} In (a), the left three columns present the results of single-view reconstruction on ShapeNet cars and chairs, while the subsequent four columns showcase fine-grained disentanglement of geometric attributes, including roofline and body straightness for cars, as well as armrest height and leg thickness for chairs. (b) demonstrates 3D disentanglement results on the visual appearance attributes including grayscale, body color and local color.}
  \vspace{-2mm}
  \label{fig:traversal_main}
\end{figure*}

\subsection{Results}
\subsubsection{Interpretable 3D Reconstruction}
To demonstrate the capability of our model in interpretable 3D reconstruction, we conduct a series of experiments across typical 3D reconstruction datasets, including ShapeNet datasets and CO3D Hydrants. As depicted in Figure~\ref{fig:traversal_main}, 3DisGS achieves fine-grained 3D disentanglement in both geometry and appearance independently, while preserving high-fidelity reconstruction. Specifically, Figure~\ref{fig:traversal_main}(a) illustrates the results of semantic disentanglement on 3D geometry, accomplished through latent traversal within the latent space. In each traversal row, a specific semantic attribute—such as rooflines, the roundness of cars and the leg thickness of chairs—varies independently, while other attributes remain unchanged. Furthermore, Figure~\ref{fig:traversal_main}(b) showcases the disentanglement of visual appearance attributes such as body color, local color and grayscale. These results show the model's capability to independently disentangle and learn meaningful semantic attributes in geometry and appearance. Additional examples are provided in the appendix.

\subsubsection{Qualitative Results}
\begin{figure}[tb]
    \centering
    \includegraphics[width=\linewidth]{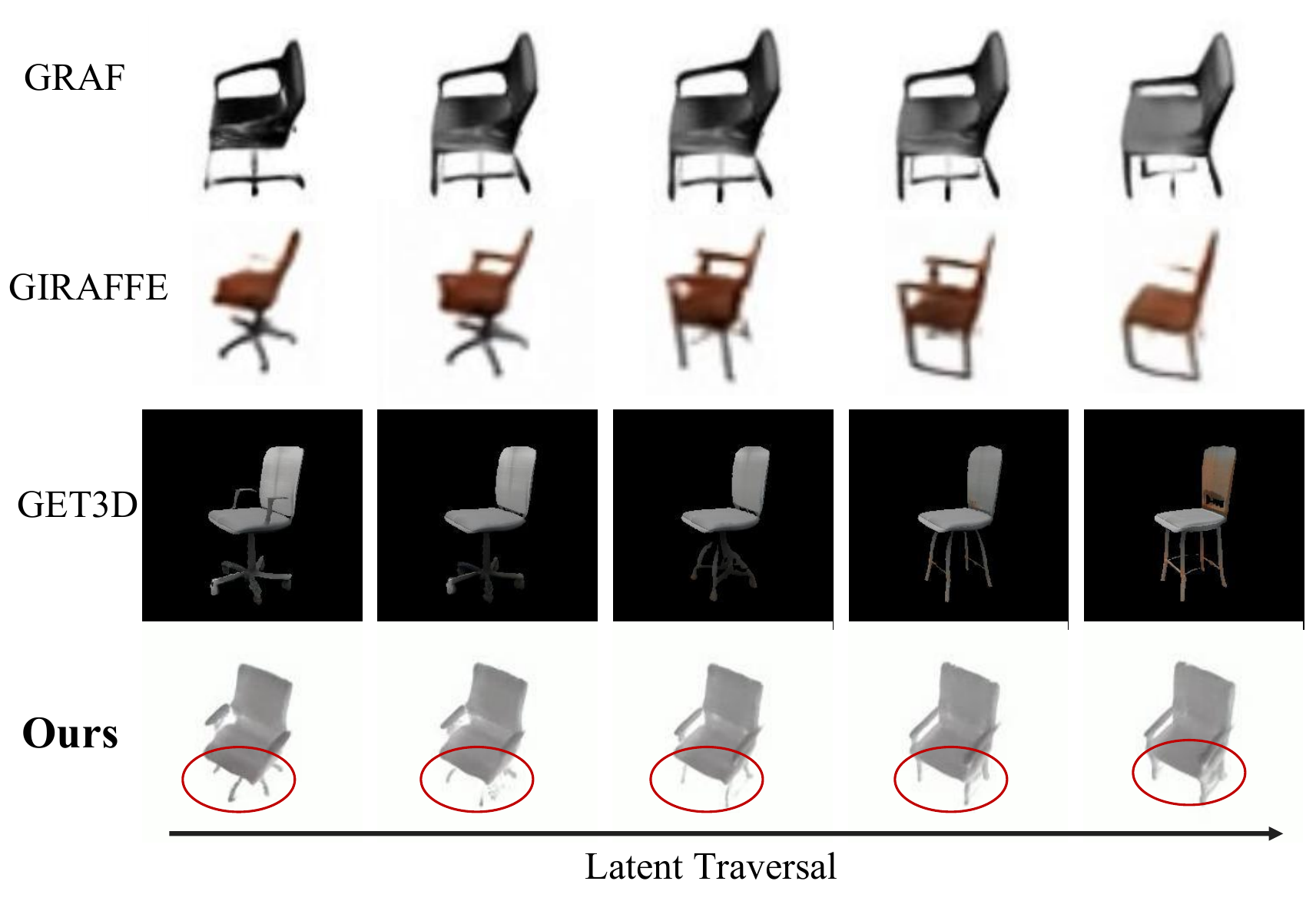}
    \caption{\textbf{Qualitative comparison results.} 3DisGS surpasses the baselines in 3D disentanglement, as it can manipulate the attributes while maintaining the integrity of irrelevant representations.}
    \label{fig:comparison}
\end{figure}

We compare 3DisGS with typical 3D-aware models that claim a certain degree of structural disentanglement, including GRAF~\cite{schwarz2020graf}, GIRAFFE~\cite{niemeyer2021giraffe}, and GET3D~\cite{gao2022get3d} on the ShapeNet dataset. Since GIRAFFE and GET3D offer pre-trained models on ShapeNet chairs, we directly utilize their checkpoints and re-train GRAF using the same dataset to ensure consistency. To illustrate their ability attribute manipulation, we perform style interpolation for them across the same attributes extracted by our model. 

As shown in Figure~\ref{fig:comparison}, we present the comparative results of continuous interpolation on attribute "chair leg style", a common attribute in chairs. The results of the baselines demonstrate that global features, such as armrest are altered simultaneously during the manipulation. Notably, even though only shape code is changed, baseline models still exhibit appearance change, such as backrest color, indicated insufficient disentanglement between shape and appearance. It shows that 3DisGS surpasses the baselines in semantic disentanglement, as it can manipulate the disentangled attributes while maintaining the integrity of irrelevant representations.

\subsubsection{Quantitative Results}
We perform quantitative evaluations to assess the reconstruction quality of 3DisGS in comparison to state-of-the-art single-view 3D reconstruction models. Specifically, we evaluate the reconstruction performance of both 3DisGS and the baseline models using the ShapeNet Chairs dataset. \cref{tab:comp} reports the results of Peak Singal-to-Noise Ratio~(PSNR), Structural Similarity~(SSIM) and LPIPS~\cite{zhang2018unreasonable} scores. Compared to typical 3D reconstruction methods, our method achieves comparable LPIPS and SSIM scores to the state-of-the-art methods and inferior PSNR scores. We attribute this reduction in PSNR as a tradeoff between interoperability and reconstruction quality. Furthermore, as demonstrated in \cref{tab:efficiency}, the proposed model exhibits competitive performance in terms of computational efficiency and convergence speed.
\begin{table}
\resizebox{\linewidth}{!}{

    \begin{tabular}{l|cccc}
        \toprule
                 & Dis.  & PSNR $\uparrow$ & LPIPS $\downarrow$ & SSIM $\uparrow$ \\
\hline
SRN\cite{sitzmann2019srns}              & \xmark & 22.89 & 0.104 & 0.89 \\

FE-NVS\cite{guo2022fast}           & \xmark & 23.21 & 0.077 & 0.92 \\

PixelNeRF\cite{yu2021pixelnerf}        & \xmark & 23.72 & 0.128 & 0.90 \\

SplatterImage\cite{szymanowicz2024splatter}    & \xmark & 24.43 & 0.067 & 0.93 \\

TriplaneGaussian\cite{zou2024triplane} & \xmark & 22.72 & 0.076 & 0.94 \\
\hline
Ours             & \cmark & 21.40 & 0.102 & 0.93 \\
        \bottomrule

    \end{tabular}
    
}
    \caption{\textbf{Quantitative comparison with state-of-the-art models.} 3DisGS demonstrates performance comparable to baselines, despite incorporating interpretability that introduces a tradeoff in quality.}
    \label{tab:comp}
\end{table}

\begin{table}
\centering
    \resizebox{\linewidth}{!}{
    \begin{tabular}{l|cccc}
\toprule
                 & Params(M)  & Mem.(G)  & TT(hrs)  & Epochs  \\
\hline
TriplaneGaussian & \textbf{102.6} & 28.8 & 70.5 & 12 \\

Ours             & 142.4 & \textbf{20.7} & \textbf{48.7} & 8 \\
\bottomrule
    \end{tabular}}
    \caption{\textbf{Quantitative comparison on computation efficiency.} We conduct comparison with the baseline in terms of parameter size, memory consumption, training time (TT) and training epochs.}
    \label{tab:efficiency}
\end{table}

\subsection{Ablation Study}
To validate the effectiveness of different components in 3DisGS, we conduct an ablations over DRL constraints, DRL-based encoder-adapters, geometry initialization and mutual information loss.\\

\noindent \textbf{w/o DRL constraints.} We compared our full model to both a baseline and a version without DRL constraints to assess their impact of disentanglement. As shown in \cref{tab:ablation-kl}, removing DRL constraints improves reconstruction quality but reduces disentanglement. Both models underperform compared to the baseline, illustrating the trade-off between reconstruction quality and disentanglement.\\

\vspace{-2mm}

\noindent \textbf{Style-guided reconstruction module.} The style-guided reconstruction module is crucial for achieving view-consistent, high-quality 3D reconstruction. To demonstrate its significance, we compared 3DisGS with baseline variants incorporating a 2D adapter for DRL and a transformer decoder conditioned on image-like features. As shown in \cref{tab:ablation-style}, the inclusion of the style-guided module significantly improves reconstruction quality, underscoring its critical contribution to the overall performance of our framework.\\

\begin{figure}
    \centering
    \includegraphics[width=\linewidth]{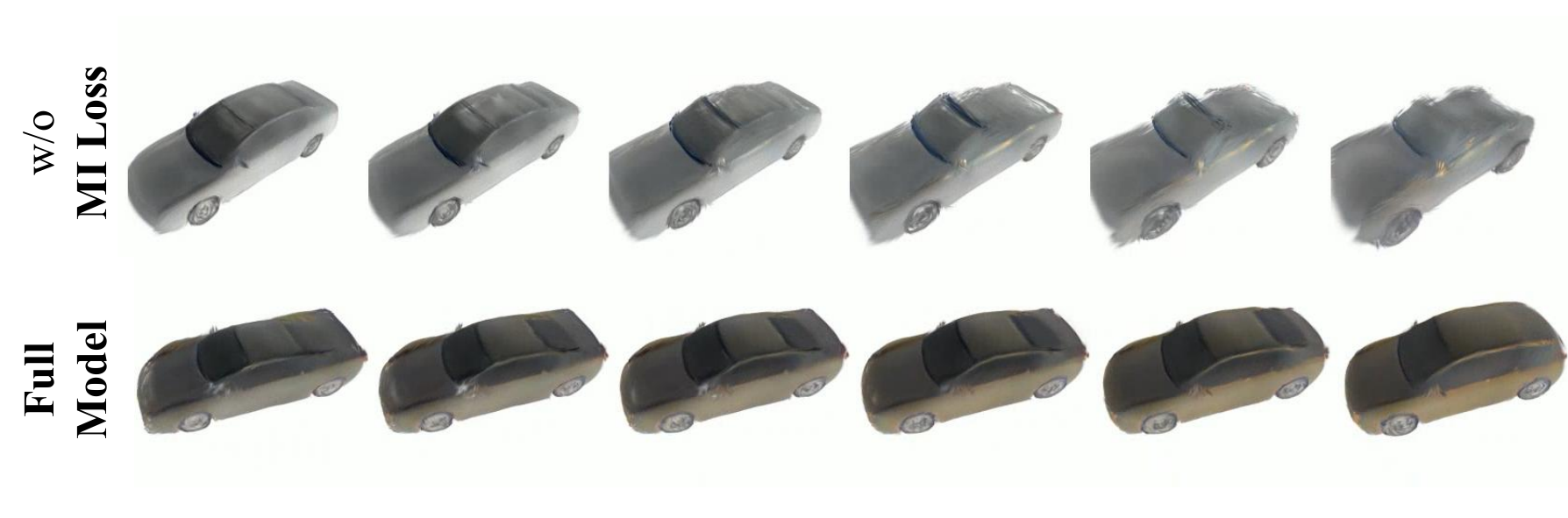}
    \vspace{-2mm}
    \caption{\textbf{Ablation study on the Mutual Information~(MI) Loss.} The absence of the MI loss leads to observable artifacts.}
    \label{fig:ablation-mi}
\end{figure}

\vspace{-2mm}
\noindent \textbf{w/o mutual information loss.} To demonstrate the importance of the mutual information loss on enforcing 3D information transformation, we conducted a comparative analysis of our model w/o the inclusion of this loss function. As demonstrated in Figure.~\ref{fig:ablation-mi}, the absence of mutual information loss results in reduced disentanglement and a tendency to overfit geometric information, ultimately leading to suboptimal performance in 3D reconstruction.

\section{Discussion}
\textbf{1) 3DGS vs. NeRF in 3D disentanglement:} From our perspective, the 3DGS framework, as an implicit-explicit hybrid approach, demonstrates greater suitability for 3D disentanglement compared to NeRF-based methods. This superiority stems from its discrete nature, which inherently enables the mapping of each Gaussian component to disentangled semantic attributes identified by the DRL models. \textbf{2) Future work:} In the next phase, we aim to enhance 3DisGS by enabling it to capture environmental variations, such as shadows, light rays, reflections and \etc by representing them as disentangled latent factors. This extension has the potential to address critical challenges in the 3D reconstruction domain, particularly in scenarios requiring accurate modeling of environmental effects. \textbf{3) Current limitations:} the primary limitation of this work lies in the trade-off between reconstruction quality and interpretability. In future iterations, we plan to address this challenge by incorporating additional modules and designing tailored loss functions.
\label{sec:limitations}
\begin{table}[tb]
    \centering
    \begin{tabular}{l|ccc}
    \toprule
                 & PSNR $\uparrow$ & LPIPS $\downarrow$ & SSIM $\uparrow$ \\
\hline
TriplaneGaussian & 17.80 & 0.18 & 0.80 \\

Ours~(w/o KL)     & 17.10 & 0.20 & 0.79 \\

Ours~(full)       & 16.61 & 0.21 & 0.78 \\
    \bottomrule
    \end{tabular}
    \caption{\textbf{Ablation study on DRL constraints.} We compare the results of baseline TriplaneGaussian model, our model without KL divergence constraints, and the full model include DRL constraints. }
    \label{tab:ablation-kl}
\end{table}
\begin{table}[tb]
    \centering
    \begin{tabular}{l|ccccc}
    \toprule
  & Sty.P.  & Sty.T.  & PSNR $\uparrow$ & LPIPS $\downarrow$ & SSIM $\uparrow$ \\
\hline
a & \xmark & \xmark & 15.06 & 0.24 & 0.75 \\

b & \xmark & \cmark & 15.52 & 0.22 & 0.76 \\

c & \cmark & \xmark & 15.75 & 0.22 & 0.76 \\

d & \cmark & \cmark & \textbf{16.61} & \textbf{0.21} & \textbf{0.78} \\
    \bottomrule
    \end{tabular}
    \caption{\textbf{Ablation study on reconstruction module designs}. It includes style-based point cloud reconstruction (Sty.P.) and style-guided triplane reconstruction (Sty.T.), evaluated against naive transformer-based variants.}
    \label{tab:ablation-style}
\end{table}
\section{Conclusion}
This work proposes a single-view interpretable 3DGS model that leverages a hierarchical DRL strategy to discover both coarse- and fine-grained 3D semantics. The dual-branch framework, comprising a point cloud initialization branch and a triplane-Gaussian generation branch, achieves coarse-grained disentanglement by separating geometry and appearance features. Subsequently, fine-grained semantic representations within each modality are further discovered via DRL-based encoder-adapters. To our knowledge, it is the first work to achieve unsupervised and interpretable 3DGS. 

\section*{Acknowledgements}
This work was supported by the National Natural Science Foundation of China (Grants 62302246), the Natural Science Foundation of Zhejiang Province, China (Grant LQ23F010008), the IDT Foundation of Youth Doctoral Innovation~(Grant S203.2.01.32.002).
Additional support was provided by the High Performance Computing Center at Eastern Institute of Technology, Ningbo, and Ningbo Institute of Digital Twin.

\label{sec:conclusion}

{\small
\bibliographystyle{unsrt}
\bibliography{references}
}

\end{document}